\documentclass[final,3p,times]{elsarticle}  

\usepackage{amsmath,amssymb}
\usepackage{hyperref}       
\usepackage{enumitem}        
\usepackage{graphicx}       
\usepackage[export]{adjustbox}  
\usepackage{subcaption}
\usepackage[font=small,labelfont=bf]{caption}
\usepackage{hyperref}
\journal{Neurocomputing}
\begin{document}
\begin{frontmatter}
\title{NM-Hebb: Coupling Local Hebbian Plasticity with Metric Learning for More Accurate and Interpretable CNNs}
\author[TttechAuto]{Davorin Miličević}
\ead{davorin.milicevic@tttech-auto.com}
\author[ferit]{Ratko Grbić\corref{cor1}}
\ead{ratko.grbic@ferit.hr}
\cortext[cor1]{Corresponding author}

\affiliation[TttechAuto]{
  organization={TTTech Auto part of NXP},
  addressline={Cara Hadrijana 10B},
  city={Osijek},
  postcode={31000},
  country={Croatia}
}

\affiliation[ferit]{organization={Faculty of Electrical Engineering, Computer Science and Information Technology Osijek},
           addressline={Kneza Trpimira 2B},
           city={Osijek},
           postcode={HR-31000},
           country={Croatia}}

\begin{abstract}
Deep Convolutional Neural Networks (CNNs) achieve high accuracy but often rely on purely global, gradient-based optimisation, which can lead to overfitting, redundant filters, and reduced interpretability. To address these limitations, we propose NM-Hebb, a two-phase training framework that integrates neuro-inspired local plasticity with distance-aware supervision. Phase 1 extends standard supervised training by jointly optimising a cross-entropy objective with two biologically inspired mechanisms: (i)~a Hebbian regulariser that aligns the spatial mean of activations with the mean of the corresponding convolutional filter weights, encouraging structured, reusable primitives; and (ii)~a learnable neuromodulator that gates an elastic-weight–style consolidation loss, preserving beneficial parameters without freezing the network. Phase~2 fine-tunes the backbone with a pairwise metric-learning loss, explicitly compressing intra-class distances and enlarging inter-class margins in the embedding space. Evaluated on CIFAR-10, CIFAR-100, and TinyImageNet across five backbones (ResNet-18, VGG-11, MobileNet-v2, EfficientNet-V2, DenseNet-121), NM-Hebb achieves consistent gains over baseline and other methods: Top-1 accuracy improves by +2.0–10.0 pp (CIFAR-10), +2.0–9.0 pp (CIFAR-100), and up to +4.3-8.9 pp (TinyImageNet), with Normalised Mutual Information (NMI) increased by up to +0.15. Qualitative visualisations and filter-level analyses further confirm that NM-Hebb produces more structured and selective features, yielding tighter and more interpretable class clusters. Overall, coupling local Hebbian plasticity with metric-based fine-tuning yields CNNs that are not only more accurate but also more interpretable, offering practical benefits for resource-constrained and safety-critical AI deployments.
\end{abstract}

\begin{highlights}
\item Two-phase CNN training with Hebbian plasticity and neuromodulation
\item +2–10 pp accuracy and +0.06–0.15 NMI gains across multiple datasets 
\item Produces tighter, more interpretable class clusters in the embedding space
\end{highlights}

\end{frontmatter}
\clearpage\phantomsection
\section{Introduction}
Convolutional Neural Networks (CNNs) have become the \emph{de facto}
standard for visual recognition, with performance steadily rising from
the early VGG\,\cite{simonyan2014very} and ResNet\,\cite{he2016deep}
families to lightweight designs such as MobileNet-v2\,%
\cite{sandler2018mobilenetv2} and EfficientNet\,\cite{tan2019efficientnet}.
More recent pure-CNN architectures show that carefully re-examined
design choices—e.g. very large depth-wise kernels in
RepLKNet\,\cite{ding2022replknet}, compound-scaling with
fused-MBConv blocks in EfficientNetV2\,\cite{tan2021efficientnetv2},
and global-response normalisation in ConvNeXt V2\,\cite{woo2023convnextv2}—
can close or even surpass the accuracy gap to vision transformers while
retaining the efficiency and favourable inductive biases of convolutions.
However, these gains often come at the cost of ever larger models and
extensive labelled datasets, which hampers deployment on memory- and
compute-constrained platforms.

Training deep networks via standard supervised learning based on cross-entropy loss and stochastic gradient-based optimisation demands high-capacity
hardware and massive annotation effort. In many practical
settings—embedded vision, mobile robotics—only limited compute, memory,
and labelled data are available. There is thus growing interest in
methods that produce compact CNNs with competitive performance, reducing
both parameter count and reliance on large supervised datasets \cite{iandola2016squeezenet,tan2021efficientnetv2,sohn2020fixmatch}.
A variety of strategies have been explored. Semi-supervised schemes
pre-train convolutional layers with unsupervised or Hebbian rules
\cite{lagani2021semi} before fine-tuning the classifier.
Knowledge-distillation transfers information from a large teacher
to a smaller student model
\cite{hinton2015distilling,touvron2021training}, while self-distillation
iteratively refines a network on its own soft targets
\cite{liu2024rsd}. Contrastive and metric-learning
losses\,\cite{khosla2020supcon,kim2020proxyanchor} reshape the embedding
space to enforce intra-class compactness and inter-class separation.
Biologically inspired mechanisms such as Hebbian
updates\,\cite{amato2019hebbian,journe2023softhebb} and neuromodulated
plasticity\,\cite{miconi2019backpropamine} have likewise improved sample
efficiency and interpretability. Yet most existing methods rely on
multi-stage pipelines, large batch-sizes or complex schedules, and few
offer a unified end-to-end solution. In this work we integrate three elements—local Hebbian plasticity,
learnable neuromodulation, and metric supervision—into a single
two-phase framework we call Neuromodulated Hebbian (NM-Hebb).  Our
main contributions are:
\begin{enumerate}[label=\arabic*.)]
 \item Convolutional Hebbian regulariser: a local penalty that
       aligns spatial means of activations and filter weights,
       stabilising early layers;
 \item Learnable neuromodulator: a lightweight MLP that
       dynamically gates Hebbian and consolidation losses, emulating
       dopamine-like control of plasticity;
 \item Pairwise metric fine-tuning: a second phase that
       enforces intra-class cohesion and inter-class separation via a
       Euclidean-margin loss while continuing Hebbian and neuromodulated
       updates.
\end{enumerate}
Evaluated on CIFAR-10, CIFAR-100, and TinyImageNet with five different backbones
(VGG-11, ResNet-18, MobileNet-v2, EfficientNet-V2, and DenseNet-121),
NM-Hebb delivers consistent Top-1 accuracy gains of
+2–10\,pp on CIFAR-10, +2–9\,pp on CIFAR-100,
and +4.3–8.9\,pp on TinyImageNet, while significantly tightening
latent clusters (+0.07–0.15 NMI) compared with standard supervised learning based on cross-entropy loss, as well as techniques such as knowledge distillation and contrastive learning. Our work will be available upon acceptance at \url{https://github.com/Davorin51/NM-Hebb}.

The remainder of this paper is organised as follows. Section~\ref{sec:related} reviews relevant literature on biologically inspired learning, neuromodulation, and metric learning in CNNs. Section~\ref{sec:method}
presents the proposed NM-Hebb framework in detail, including the Hebbian regulariser, learnable neuromodulator, and metric fine-tuning phase. Section~\ref{sec:exp} describes the experimental setup and implementation details.
Section~\ref{sec:results} reports and analyses results and
Section~\ref{sec:concl} concludes the paper and gives potential directions for future research.

\section{Related Work}\label{sec:related}
In this section, we review four research streams most relevant to our
NM-Hebb framework, highlighting both advances and the remaining
challenges that motivate our unified approach.
\subsection{Hebbian Learning in Deep Networks}
The principle of Hebbian synaptic plasticity—``cells that fire together, wire
together''—has recently been incorporated into deep CNNs.
Amato \textit{et al.}~\cite{amato2019hebbian} introduced a local weight
update for shared convolutional kernels, showing competitive accuracy on
CIFAR-10 dataset. Lagani \textit{et al.}~\cite{lagani2021semi} extended this to a
semi-supervised setting by pre-training all convolutional layers via
Hebbian updates before fine-tuning only the final classifier.
Their FastHebb algorithm~\cite{lagani2022fasthebb} further addressed
scalability by fusing batch-level Hebbian computations and leveraging
GPU matrix operations to train on ImageNet. Journé \textit{et al.}'s
SoftHebb~\cite{journe2023softhebb} even eliminates feedback signals
entirely, attaining Top-1 accuracy of 80\% on CIFAR-10 purely with local rules.
More recently, Jiménez Nimmo and Mondragón~\cite{jimenez-nimmo2025advancing}
combined Hebbian plasticity with winner-take-all competition, lateral
inhibition and the BCM rule to match back-prop baselines while yielding
sparse, hierarchical features.
Despite these successes, existing Hebbian schemes typically require
multi-stage pipelines, careful balancing of plasticity coefficients and
only recently scale beyond small datasets. Our work addresses these limitations by integrating Hebbian updates into both single-image and pairwise training phases, with plasticity dynamically modulated by a learnable neuromodulator.
\subsection{Neuromodulation and Synaptic Consolidation}
In biological systems, plasticity is gated by neuromodulators such as
dopamine. Differentiable Plasticity~\cite{miconi2018diffplastic}
assigns each connection a learnable Hebbian coefficient optimised by
back-propagation; Backpropamine~\cite{miconi2019backpropamine} adds an MLP that
dynamically scales plasticity based on the loss. In the spiking domain,
SSTDP~\cite{liu2021sstdp} combines global error signals with local
spike-timing rules. Continual-learning methods such as EWC
\cite{kirkpatrick2017overcoming} and Synaptic Intelligence
\cite{zenke2017si} emulate dopamine-like consolidation to mitigate
catastrophic forgetting.
Most such schemes target multi-task settings or rely on per-synapse
coefficients, increasing complexity. In contrast, we embed one
lightweight neuromodulator inside a single-task pipeline, gating both
Hebbian and consolidation losses to obtain a favourable
plasticity–stability trade-off with minimal overhead. Several optimiser-side regularisers have also been proposed.
Sharpness-Aware Minimisation (SAM) \cite{foret2021sam} explicitly
penalises local curvature and improves generalisation across
architectures. Unlike NM-Hebb, however, SAM operates in weight space and
does not explicitly shape embedding geometry.
\subsection{Metric Learning for Intra-Class Structure}
Metric losses explicitly shape the geometry of embedding space.
ArcFace~\cite{deng2019arcface}, SoftTriple~\cite{qian2019softtriple} and
Proxy-Anchor~\cite{kim2020proxyanchor} introduce angular or proxy-based
margins, while Supervised Contrastive (SupCon)~\cite{khosla2020supcon}
pulls all same-class samples together and pushes different-class ones
apart. Latent Boost~\cite{geissler2024latentboost} optimises latent
compactness directly; Opitz and Ropinski~\cite{opitz2024embedding}
provide a systematic comparison of embedding geometries.
These methods yield tight clusters but often depend on large batches or
complex mining. In contrast our pairwise phase uses a simple margin loss, combined
with Hebbian regularisation and neuromodulation, to produce compact
intra-class clusters and robust inter-class separation without the need for large
batch sizes.
\subsection{Distillation and Self-Distillation}
Knowledge distillation transfers a teacher’s softened outputs to a
student network, improving generalisation
\cite{hinton2015distilling,touvron2021training}. Repeated
Self-Distillation (RSD)~\cite{liu2024rsd} iteratively refines a model on
its own predictions. These approaches boost accuracy and enable
compression, yet require a large teacher or plateau after a few rounds and
do not explicitly control embedding compactness.
NM-Hebb complements distillation by directly shaping the embedding space
through metric supervision and biologically inspired plasticity,
delivering higher accuracy and interpretability across CNN architectures
without external teachers or iterative self-labelling.

\section{Methodology}\label{sec:method}
\subsection{Overall two-phase training framework}
NM-Hebb learns in two sequential stages. Phase 1 minimises a combined loss consisting of cross-entropy and a neuromodulated local Hebbian regulariser, the latter aligning filter activations with their weights to stabilise early convolutional features:
       \begin{equation}
       \label{eq:phase1_full}
       \mathcal L^{(1)}(\theta,\phi) =
            L_{\mathrm{CE}}
            + \lambda_{\mathrm{hebb}}^{(1)}
              \,\nu_\phi\!\bigl(L_{\mathrm{CE}}\bigr)
              \,\mathcal R_{\mathrm{Hebb}}(x),
       \end{equation}
       where:
       \begin{itemize}
         \item $L_{\mathrm{CE}}$ – standard cross-entropy loss,
         \item $\nu_\phi$ – neuromodulatory gate $\in(0,1)$ from an MLP,
         \item $\lambda_{\mathrm{hebb}}^{(1)}$ – scalar weighting of the Hebbian term,
         \item $\mathcal R_{\mathrm{Hebb}}$ – Hebbian regulariser aligning mean activations and weights,
         \item $x$ – input sample and $y$ its corresponding label,
         \item $\theta$ – parameters of the backbone network, and
         \item $\phi$ – parameters of the neuromodulatory MLP controlling the regularisation strength.
       \end{itemize}
       The Hebbian regulariser $\mathcal R_{\mathrm{Hebb}}$ is applied to a single convolutional layer per backbone chosen in the early-to-mid hierarchy 
to stabilise structural features. After optimisation over all mini-batches in Phase~1,
       the backbone parameters are copied to a frozen set
       $\theta^{(-)}$ for use in Phase~2.

Phase 2 fine-tunes the backbone with a pairwise metric loss, reducing intra-class variance and enlarging inter-class margins in the embedding space:
\begin{equation}
\label{eq:phase2_full}
\mathcal L^{(2)}(\theta,\phi) =
    L_{\mathrm{CE},A} + L_{\mathrm{CE},B}
  + \lambda_{\mathrm{metric}}\,L_{\mathrm{metric}}
  + \nu_\phi\!\left(
      \lambda_{\mathrm{cons}}\,{\lVert\theta - \theta^{(-)}\*\rVert}_{2}^{2}
    + \lambda_{\mathrm{hebb}}^{(2)}
      \,\tfrac12\left[
        \mathcal R_{\mathrm{Hebb}}(x_A)
      + \mathcal R_{\mathrm{Hebb}}(x_B)
      \right]
    \right),
\end{equation}
       where:
       \begin{itemize}
         \item $L_{\mathrm{CE},A}, L_{\mathrm{CE},B}$ – cross-entropy losses for samples A and B,
         \item $L_{\mathrm{metric}}$ – pairwise metric loss promoting intra-class compactness and inter-class separation,
         \item $\lambda_{\mathrm{metric}}$ – weighting of the metric term,
         \item $\lambda_{\mathrm{cons}}$ – weighting of the consolidation term and
         \item $\theta^{(-)}$ – backbone weights saved from Phase~1.
       \end{itemize}

Both parameter sets $(\theta,\phi)$ are updated by in
each stage; only the objective terms change.

\subsection{Phase 1: single-image training}
Phase~1 performs fully supervised learning on individual labelled
images $(x,y)$. The aim is to establish decision boundaries with the
cross-entropy loss while a biologically motivated Hebbian term stabilises
local features. For each mini-batch, the loss \eqref{eq:phase1_full} is minimised 
using stochastic gradient descent with Nesterov momentum, with
gradients computed by standard supervised learning based on cross-entropy loss.

Cross-entropy term $L_{\mathrm{CE}}$ is defined as:
\begin{equation}
\label{eq:cross_entropy}
 L_{\mathrm{CE}}(x,y)=
 -\sum_{c=1}^{K}\mathbf 1_{[y=c]}\,\log p_{\theta}(c\mid x),
\end{equation}
where $p_{\theta}(c\mid x)$ is the soft-max output. Minimising
$L_{\mathrm{CE}}$ compels the network to assign high probability to the
true class.

The Hebbian regulariser $\mathcal R_{\mathrm{Hebb}}$ in 
Equation~\eqref{eq:hebb_reg} encourages stable and consistent 
feature extraction by penalising the squared difference between the 
mean activation $\bar{a}_f$ and the mean kernel weight $\bar{w}_f$ for 
each output channel $f$. This alignment helps maintain coherent filter 
statistics throughout training, reducing drift in early convolutional layers:
\begin{equation}
\label{eq:hebb_reg}
\mathcal R_{\mathrm{Hebb}}(x)=
\frac{1}{C_{\mathrm{out}}}
\sum_{f=1}^{C_{\mathrm{out}}}
\bigl(\bar a_f - \bar w_f\bigr)^{2},
\quad
\begin{aligned}
\bar a_f &= \frac{1}{N H W}
          \sum_{i=1}^{N}\sum_{u=1}^{H}\sum_{v=1}^{W}
          a_{\max}^{(i)}[f,u,v],\\
\bar w_f &= \frac{1}{C_{\mathrm{in}} K^{2}}
          \sum_{c=1}^{C_{\mathrm{in}}}\sum_{u=1}^{K}\sum_{v=1}^{K}
          w_{f}[c,u,v].
\end{aligned}
\end{equation}
Here $C_{\mathrm{out}}$/$C_{\mathrm{in}}$ are the output/input channel
counts, $N$ the mini-batch size, $H{\times}W$ the spatial resolution,
and $K{\times}K$ the kernel size.

Neuromodulator $\nu_{\phi}$ is used to adaptively gate the regularisation strength based on the 
current classification loss:
\begin{equation}
\label{eq:neuromodulator}
 \nu_{\phi}(L_{\mathrm{CE}})=
 \sigma\!\bigl(\mathrm{MLP}_{\phi}(L_{\mathrm{CE}})\bigr),
\end{equation}
where $\phi$ are the parameters of the neuromodulatory MLP and 
$\sigma(\cdot)$ denotes the sigmoid activation function. 
The high classification error increases the gate, allowing stronger Hebbian consolidation; as the error decreases, the gate relaxes, preventing overconstraint.
The neuromodulator is a lightweight two-layer MLP that maps 
the scalar cross-entropy loss to a coefficient in $(0,1)$. It consists of 
a hidden layer with 8 ReLU-activated units and output layer with a single neuron with sigmoid activation function. This adds fewer than 100 parameters, 
providing adaptive gating with negligible overhead.

Balancing coefficient $\lambda_{\mathrm{hebb}}^{(1)}$ is a
scalar that controls the relative strength of the Hebbian term in Phase 1 and is
set via validation search. With this combination, Phase~1 learns both what to classify
(via $L_{\mathrm{CE}}$) and how to organise early
representations (via $\mathcal R_{\mathrm{Hebb}}$ gated by
$\nu_{\phi}$), laying a stable foundation for the pairwise fine-tuning
in Phase~2.

\subsection{Phase 2: pairwise fine-tuning}
Phase~2 refines the embedding space on image pairs while preserving the
knowledge acquired in Phase~1. The procedure is as follows:
\begin{enumerate}[label=\arabic*.)]
 \item Draw pairs $\bigl((x_A,y_A),(x_B,y_B)\bigr)$ with equal
       probability of same-class or different-class membership.
 \item Compute embeddings
       $e_A=\mathrm{emb}_{\theta}(x_A)$ and
       $e_B=\mathrm{emb}_{\theta}(x_B)$ from the penultimate layer.
 \item Optimise the combined loss
       $\mathcal L^{(2)}(\theta,\phi)$ in
       Eq.~\eqref{eq:phase2_full} with respect to both $\theta$ and the
       neuromodulator parameters $\phi$.
\end{enumerate}
\begin{equation}
\label{eq:phase2_full_details}
\begin{aligned}
\mathcal L^{(2)}(\theta,\phi)=\;
&\underbrace{L_{\mathrm{CE}}(x_A,y_A)+L_{\mathrm{CE}}(x_B,y_B)}_{\text{classification}} \\[2pt]
&+\,\lambda_{\mathrm{metric}}\,
    \underbrace{L_{\mathrm{metric}}(e_A,e_B)}_{\text{embedding shaping}} \\[2pt]
&+\,\underbrace{\nu_\phi\,\lambda_{\mathrm{cons}}\,\lVert\theta-\theta^{(-)}\rVert_2^2}_{\text{consolidation}} \\[2pt]
&+\,\underbrace{\nu_\phi\,\lambda_{\mathrm{hebb}}^{(2)}\,
      \tfrac12\bigl[\mathcal R_{\mathrm{Hebb}}(x_A)+\mathcal R_{\mathrm{Hebb}}(x_B)\bigr]}_{\text{continued Hebbian}}.
\end{aligned}
\end{equation}

\begin{itemize}[itemsep=0.5em]
  \item $L_{\mathrm{CE}}(x,y)$ – cross-entropy loss on a single image $(x,y)$, ensuring that classification accuracy is maintained while the embedding space is refined.
  
  \item $L_{\mathrm{metric}}(e_A,e_B)$ – euclidean-margin loss that pulls same-class pairs together and pushes different-class pairs apart by a margin~$m$:
        \[
          L_{\mathrm{metric}}(e_A,e_B)=
          \begin{cases}
            \lVert e_A-e_B\rVert_2^2, & y_A=y_B,\\[4pt]
            \bigl[\max(0,m-\lVert e_A-e_B\rVert_2)\bigr]^2, & y_A\neq y_B.
          \end{cases}
        \]
  
  \item $\lVert\theta-\theta^{(-)}\rVert_2^2$ – quadratic penalty on deviation from Phase~1 parameters, gated by $\nu_\phi$ to emulate synaptic consolidation.
  
  \item $\mathcal R_{\mathrm{Hebb}}(x_{A/B})$ – Hebbian regularisation applied to both samples, reinforcing local feature stability during fine-tuning.
  
  \item $\nu_\phi(\cdot)$ – neuromodulator that adaptively gates consolidation and Hebbian terms according to the current classification error.
  
  \item $\lambda_{\mathrm{metric}},\lambda_{\mathrm{cons}},\lambda_{\mathrm{hebb}}^{(2)}$ – hyperparameters controlling the strength of metric shaping, consolidation, and Hebbian regularisation.
\end{itemize}

As in Phase~1, both $\theta$ and $\phi$ are updated end-to-end via standard supervised learning, resulting in embeddings that form tight intra-class clusters, while filters are consolidated and classification performance remains strong.
\section{Experimental Setup}\label{sec:exp}

\subsection{Baseline and comparative methods}
We treat standard supervised CNN training as the sole baseline, instantiated on five backbones:
VGG-11~\cite{simonyan2014very}, ResNet-18~\cite{he2016deep}, MobileNet-v2~\cite{sandler2018mobilenetv2}, EfficientNet-V2~\cite{tan2021efficientnetv2}, and DenseNet-121~\cite{huang2017densenet}.
All other families below are comparative methods chosen to span the main paradigms relevant to NM-Hebb: distillation-based training, metric-learning frameworks, and self-distillation.

\begin{itemize}
  \item Knowledge distillation (KD) – we use the teacher–student framework from Hinton \emph{et~al.}~\cite{hinton2015distilling} as a reference global-to-local signal contrast to NM-Hebb’s local neuromodulation.
  \item Self-distillation – we compare against Repeated Self-Distillation (RSD)~\cite{liu2024rsd}, which iteratively fine-tunes a model on its own predictions, analogous to Phase~2 of NM-Hebb without the Hebbian term.
  \item Metric learning – we include Supervised Contrastive Learning (SupCon)~\cite{khosla2020supcon}, which uses contrastive loss on positive/negative pairs to shape the embedding space; NM-Hebb extends this principle with neuromodulation and consolidation.
\end{itemize}

These categories were chosen because they represent the main paradigms relevant to our method: standard supervised learning, distillation-based methods, metric-learning frameworks. Together, they allow us to benchmark NM-Hebb across a broad spectrum of techniques that emphasise different combinations of performance, representation learning, and stability.
For all methods, we used the official code repositories provided by the original authors wherever available, with minimal adaptations to our environment (PyTorch 2.1, CUDA 11.7). Table~\ref{tab:baselines} summarises the baseline (top) and the comparative method families and how they relate to NM-Hebb.

\begin{table}[htbp]
\centering
\caption{Baseline and method families for comparison with NM-Hebb.}
\label{tab:baselines}
\small
\begin{tabular}{|p{3.8cm}|p{6.4cm}|p{4.1cm}|}
\hline
Category & Representative & Role in comparison \\
\hline
\multicolumn{3}{|c|}{Baseline} \\
\hline
Standard supervised CNN &
VGG-11~\cite{simonyan2014very},
ResNet-18~\cite{he2016deep},
MobileNet-v2~\cite{sandler2018mobilenetv2},
EfficientNet-V2~\cite{tan2021efficientnetv2},
DenseNet-121~\cite{huang2017densenet} &
Reference trained with standard supervised learning based on cross-entropy loss; no Hebbian term, no metric loss. \\
\hline
\multicolumn{3}{|c|}{Comparative methods} \\
\hline
Knowledge distillation (KD) & Teacher$\rightarrow$student KD~\cite{hinton2015distilling,touvron2021training} & Uses global teacher (Resnet50). \\
\hline
Self-distillation & Repeated self-distillation (RSD)~\cite{liu2024rsd} & Fine-tuning without external teacher; analogous to Phase~2 without Hebbian term. \\
\hline
Siamese / triplet & SupCon~\cite{khosla2020supcon} & Shares pairwise metric loss but lacks Hebbian regulariser. \\
\hline
\end{tabular}
\end{table}
\subsection{Architectures}
Five CNN architectures were used without any structural modifications. Hebbian regularisation is applied to a single chosen convolutional layer in each network, and the embedding vector is taken from the penultimate layer. Table~\ref{tab:arch} lists details of the backbones, including the number of parameters, the Hebbian layer, and the embedding dimension.
\begin{table}[ht]
\caption{Backbone architectures with parameter count, Hebbian regulariser location and embedding dimensionality.}
\label{tab:arch}
\centering
\begin{tabular}{|l|c|c|c|}
\hline
Backbone & Parameters (M) & Hebbian regulariser location & Embedding dimension \\
\hline
VGG-11 & 10.8 & Block~1, conv2 & 128 \\
ResNet-18 & 11.2 & Layer~2, final 3$\times$3 conv & 512 \\
MobileNet-v2 & 3.5 & Last depthwise conv & 1280 \\
EfficientNet-V2 & 5.3 & MBConv6 in Stage~3 & 128 \\
DenseNet-121 & 8.0 & Last 3$\times$3 conv in dense block~3 & 1024 \\
\hline
\end{tabular}
\end{table}

\subsection{Datasets and preprocessing}
All models were trained from scratch without pretraining on ImageNet or other datasets. Experiments were conducted on three standard datasets:
\begin{itemize}
  \item CIFAR-10~\cite{cifar10-dataset} – 50{,}000 training and 10{,}000 test images of 32$\times$32 pixels across 10 classes. For model selection, we split the original training set into 40{,}000 training and 10{,}000 validation samples, stratified by class.
  \item CIFAR-100~\cite{cifar100-dataset} – 50{,}000 training and 10{,}000 test images of 32$\times$32 pixels across 100 classes. For model selection, we split the original training set into 40{,}000 training and 10{,}000 validation samples, stratified by class.
    \item Tiny ImageNet~\cite{tiny-imagenet-200} – 100{,}000 training and 10{,}000 validation images of 64$\times$64 pixels across 200 classes. The validation set is used exclusively for model selection. For evaluation, we additionally use a community-provided test set (10{,}000 images with reconstructed labels) \cite{tiny-imagenet-unofficial}, which supplies ground-truth annotations for the original Tiny ImageNet test split. Results are consistent across validation and unofficial test sets.

\end{itemize}
Input images were normalised per channel using dataset statistics. Applied augmentations are detailed in \ref{sec:protocol}.

\subsection{Training procedure}
\label{sec:protocol}
The same data augmentations and training schedules were used for all models:
\begin{itemize}
  \item Augmentations included AutoAugment with the CIFAR-10 policy, random horizontal flips with probability 0.5, and random cropping for CIFAR datasets or random resized crops for TinyImageNet.
  \item Phase 1 consisted of 50 epochs for CIFAR datasets and 60 epochs for TinyImageNet. We used stochastic gradient descent with Nesterov momentum 0.9, an initial learning rate of $10^{-3}$, weight decay of $10^{-5}$, a cosine learning-rate schedule, stochastic weight averaging (SWA) starting at epoch 40 for CIFAR and epoch 45 for TinyImageNet, and a batch size of 128 for CIFAR and 64 for TinyImageNet.
  \item Phase 2 also used 50 epochs for CIFAR and 60 epochs for TinyImageNet, with an initial learning rate of $10^{-4}$ and the same cosine schedule and weight decay as in Phase 1; batch sizes remained unchanged.
\end{itemize}
Validation Top-1 accuracy was recorded after each epoch, and the checkpoint with the highest accuracy was used for Phase 2. Early stopping was applied with a patience of 15 epochs, though SWA often triggered training termination earlier.

\subsection{Implementation details}
All experiments were run on a single NVIDIA RTX 4090 GPU with 24 GB of memory. We used PyTorch 2.1 and CUDA 11.7. Random number generators for Python, NumPy, PyTorch, and CUDA were seeded to ensure reproducibility. Reported results are averages over five random initialisations. For the baseline family and other methods for comparison, we used the official reference implementations, adapting them only as necessary to integrate with our training pipeline.

\subsection{Evaluation metrics}
We used the following metrics to evaluate model performance:
\begin{itemize}
  \item Classification performance was measured using Top-1 accuracy.
  \item Cluster quality in the embedding space was quantified using Normalised Mutual Information (NMI).
  \item Qualitative assessment of separability was done through t-SNE visualisations of the embedding space.
  \item Filter selectivity and interpretability were evaluated using HAF, speckle rate and a Network Dissection probe.
\end{itemize}

\section{Results and discussion}\label{sec:results}
\subsection{Quantitative results}
We report final performance on the CIFAR-10, 100 and TinyImageNet test sets
using the checkpoint with highest validation
accuracy (see Section~\ref{sec:protocol}).
Results are shown separately for each dataset in Tables~\ref{tab:cifar10_results}-\ref{tab:tiny_results}. Metrics are Top-1 accuracy (\%) and Normalised Mutual Information (NMI). 
\begin{table}[ht]
\caption{Top-1 accuracy (\%) / NMI on CIFAR-10.}
\label{tab:cifar10_results}
\small
\centering
\begin{tabular}{|l|c|c|c|c|c|}
\hline
\textbf{Model} & \textbf{Baseline} & \textbf{KD} & \textbf{SupCon} & \textbf{Self-Distill} & \textbf{NM-Hebb} \\
\hline
ResNet-18       & 88.7 / 0.49 & 90.2 / 0.52 & 89.8 / 0.51 & 90.0 / 0.52 & 93.7 / 0.61 \\
VGG-11          & 85.2 / 0.51 & 87.0 / 0.53 & 86.8 / 0.50 & 86.9 / 0.51 & 91.8 / 0.57 \\
MobileNet-v2    & 86.3 / 0.38 & 88.0 / 0.41 & 87.5 / 0.40 & 87.8 / 0.41 & 92.1 / 0.58 \\
EfficientNet-V2 & 91.5 / 0.51 & 95.0 / 0.53 & 94.8 / 0.52 & 94.9 / 0.52 & 95.8 / 0.62 \\
DenseNet-121    & 94.4 / 0.50 & 95.0 / 0.52 & 95.1 / 0.51 & 95.2 / 0.51 & 96.4 / 0.56 \\
\hline
\end{tabular}
\end{table}

\begin{table}[ht]
\caption{Top-1 accuracy (\%) / NMI on CIFAR-100.}
\label{tab:cifar100_results}
\small
\centering
\begin{tabular}{|l|c|c|c|c|c|}
\hline
\textbf{Model} & \textbf{Baseline} & \textbf{KD} & \textbf{SupCon} & \textbf{Self-Distill} & \textbf{NM-Hebb} \\
\hline
ResNet-18       & 66.1 / 0.39 & 68.0 / 0.40 & 67.0 / 0.39 & 67.5 / 0.40 & 74.9 / 0.48 \\
VGG-11          & 65.0 / 0.41 & 67.2 / 0.42 & 66.5 / 0.41 & 66.8 / 0.41 & 71.9 / 0.44 \\
MobileNet-v2    & 64.6 / 0.40 & 66.0 / 0.42 & 65.5 / 0.41 & 65.8 / 0.42 & 73.8 / 0.45 \\
EfficientNet-V2 & 72.0 / 0.45 & 73.5 / 0.46 & 73.0 / 0.45 & 73.2 / 0.45 & 76.8 / 0.46 \\
DenseNet-121    & 75.2 / 0.40 & 76.0 / 0.41 & 75.5 / 0.40 & 75.7 / 0.40 & 78.9 / 0.42 \\
\hline
\end{tabular}
\end{table}

\begin{table}[ht]
\caption{Top-1 accuracy (\%) / NMI on TinyImageNet.}
\label{tab:tiny_results}
\small
\centering
\begin{tabular}{|l|c|c|c|c|c|}
\hline
\textbf{Model} & \textbf{Baseline} & \textbf{KD} & \textbf{SupCon} & \textbf{Self-Distill} & \textbf{NM-Hebb} \\
\hline
ResNet-18       & 51.00 / 0.30 & 53.0 / 0.31 & 52.5 / 0.31 & 52.8 / 0.31 & 59.97 / 0.36 \\
VGG-11          & 50.58 / 0.29 & 52.0 / 0.30 & 51.5 / 0.30 & 51.8 / 0.30 & 55.68 / 0.34 \\
MobileNet-v2    & 44.00 / 0.28 & 46.0 / 0.29 & 45.5 / 0.29 & 45.8 / 0.29 & 48.33 / 0.30 \\
EfficientNet-V2 & 45.00 / 0.31 & 47.5 / 0.33 & 47.0 / 0.32 & 47.3 / 0.32 & 51.02 / 0.34 \\
DenseNet-121    & 57.00 / 0.33 & 59.0 / 0.35 & 58.5 / 0.34 & 58.8 / 0.34 & 62.74 / 0.39 \\
\hline
\end{tabular}
\end{table}

\smallskip
Across all five backbones, NM-Hebb consistently achieves the highest Top-1 accuracy and the best (or tied-best) NMI compared to the baseline and all comparative methods. The performance gain is evident on all three datasets.

\textbf{Medium-capacity models (ResNet-18, VGG-11).} Compared to the baseline, NM-Hebb raises CIFAR-10 accuracy by +5.0\,pp (ResNet-18) and +6.6\,pp (VGG-11); for CIFAR-100 the gains rise to +8.8\,pp and +6.9\,pp, respectively. On TinyImageNet the gains are +9.0\,pp (51.0\% $\rightarrow$ 60.0\%) for ResNet-18 and +5.1\,pp (50.6\% $\rightarrow$ 55.7\%) for VGG-11. Teacher-student knowledge distillation narrows, but does not close—the gap, while supervised contrastive learning (SupCon) improves NMI yet lags behind NM-Hebb in accuracy, confirming the benefit of combining metric separation with Hebbian compression in a single pipeline.

\textbf{Low-capacity MobileNet-v2.} With the smallest backbone (3.5\,M parameters), we observe the largest relative gains on the CIFAR benchmarks (+5.8\,pp on CIFAR-10, +9.2\,pp on CIFAR-100) and a consistent improvement on TinyImageNet of +4.3\,pp (44.0\% $\rightarrow$ 48.3\%). This suggests that the Hebbian penalty acts as an effective structural prior, compensating for limited width by encouraging specialised, non-redundant filters (Section~\ref{sec:filters}). Notably, MobileNet-v2\,+\,NM-Hebb surpasses ResNet-18\,+\,BP across datasets while retaining approximately 3$\times$ fewer parameters.

\textbf{High-capacity EfficientNet-V2 and DenseNet-121}. For the baseline, CIFAR-10 accuracy gains with NM-Hebb are +4.3\,pp and +2.0\,pp, while on CIFAR-100 the gains are +4.8\,pp and +3.7\,pp, respectively. On TinyImageNet both backbones benefit markedly: +6.0\,pp for EfficientNet-V2 (45.0\% $\rightarrow$ 51.0\%) and +5.7\,pp for DenseNet-121 (57.0\% $\rightarrow$ 62.7\%). NMI increases across all three datasets, implying that NM-Hebb reshapes the embedding space even when cross-entropy is near saturation, and that the advantage grows with label granularity and input resolution.

Cluster quality improves especially in Phase~2, which provides the bulk of the NMI improvement across datasets while adding only $\sim$20\% extra wall-clock time. Hebbian alignment compresses intra-class variance, and the pairwise loss—guarded by neuromodulated consolidation—pushes classes apart without destabilising earlier layers. Additional visualizations can be found in Section~\ref{sec:embedding}.

The overall pattern shows that gains are inversely correlated with parameter count (strongest on MobileNet-v2), increase with label granularity (larger on CIFAR-100 and TinyImageNet), and remain consistently positive across plain, residual, inverted-bottleneck, compound-scaled, and dense designs—suggesting NM-Hebb is architecture-agnostic and particularly valuable for lightweight models and fine-grained or high-class-count tasks.

\subsection{Embedding space visualization} \label{sec:embedding}
Although the previous section quantified cluster compactness with NMI and Top-1 accuracy results,
numerical scores alone do not reveal where in the feature space
the gains arise. We therefore visualise t-SNE projections of the ResNet-18 final-layer
embeddings after Phase~1 and after the complete two-phase NM-Hebb pipeline,
for CIFAR-10 and CIFAR-100.

\begin{figure*}[htp]
  \centering
  \begin{subfigure}[t]{0.46\textwidth}
    \fbox{\includegraphics[width=\linewidth]{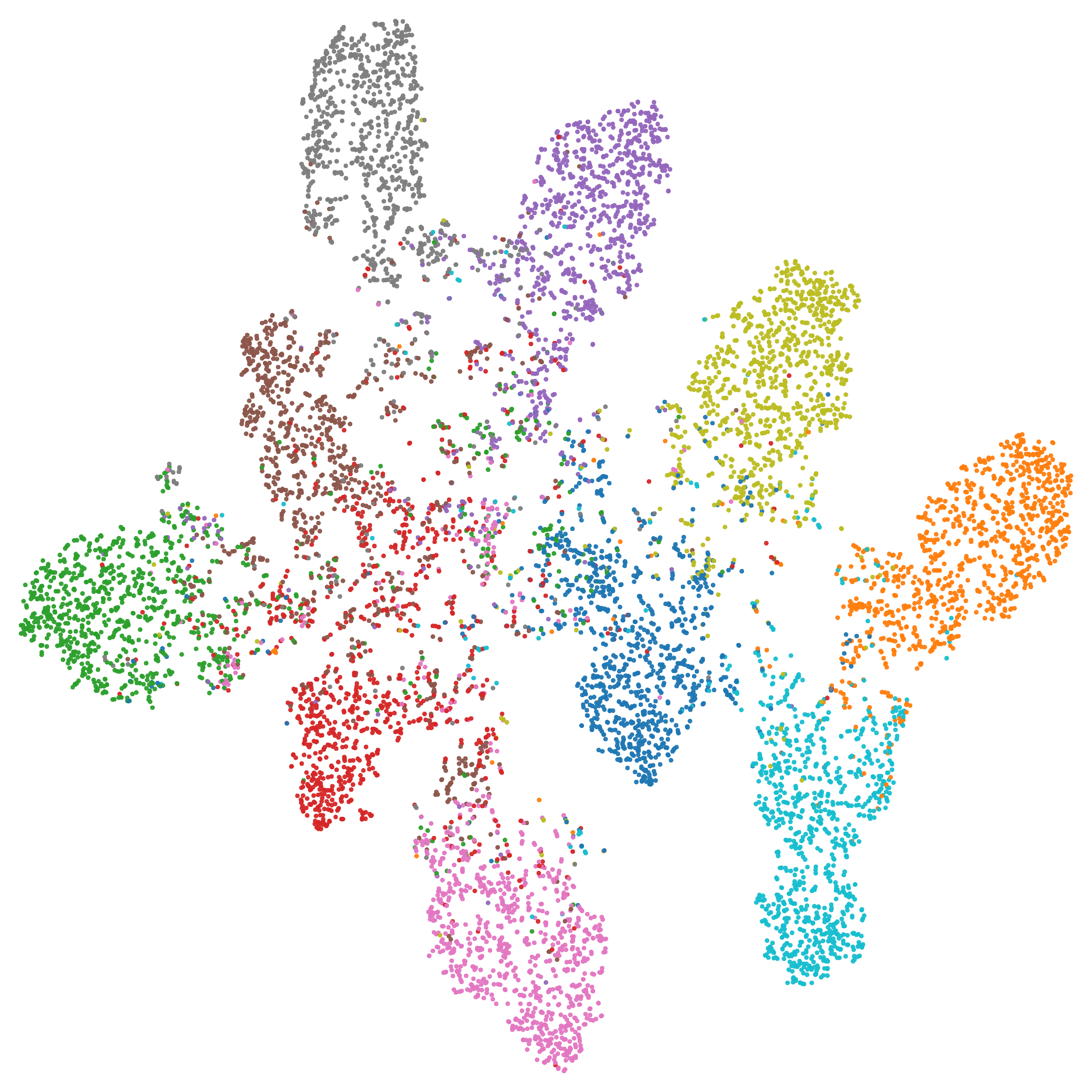}}
    \caption{CIFAR-10, Phase 1}
  \end{subfigure}\hfill
  \begin{subfigure}[t]{0.46\textwidth}
    \fbox{\includegraphics[width=\linewidth]{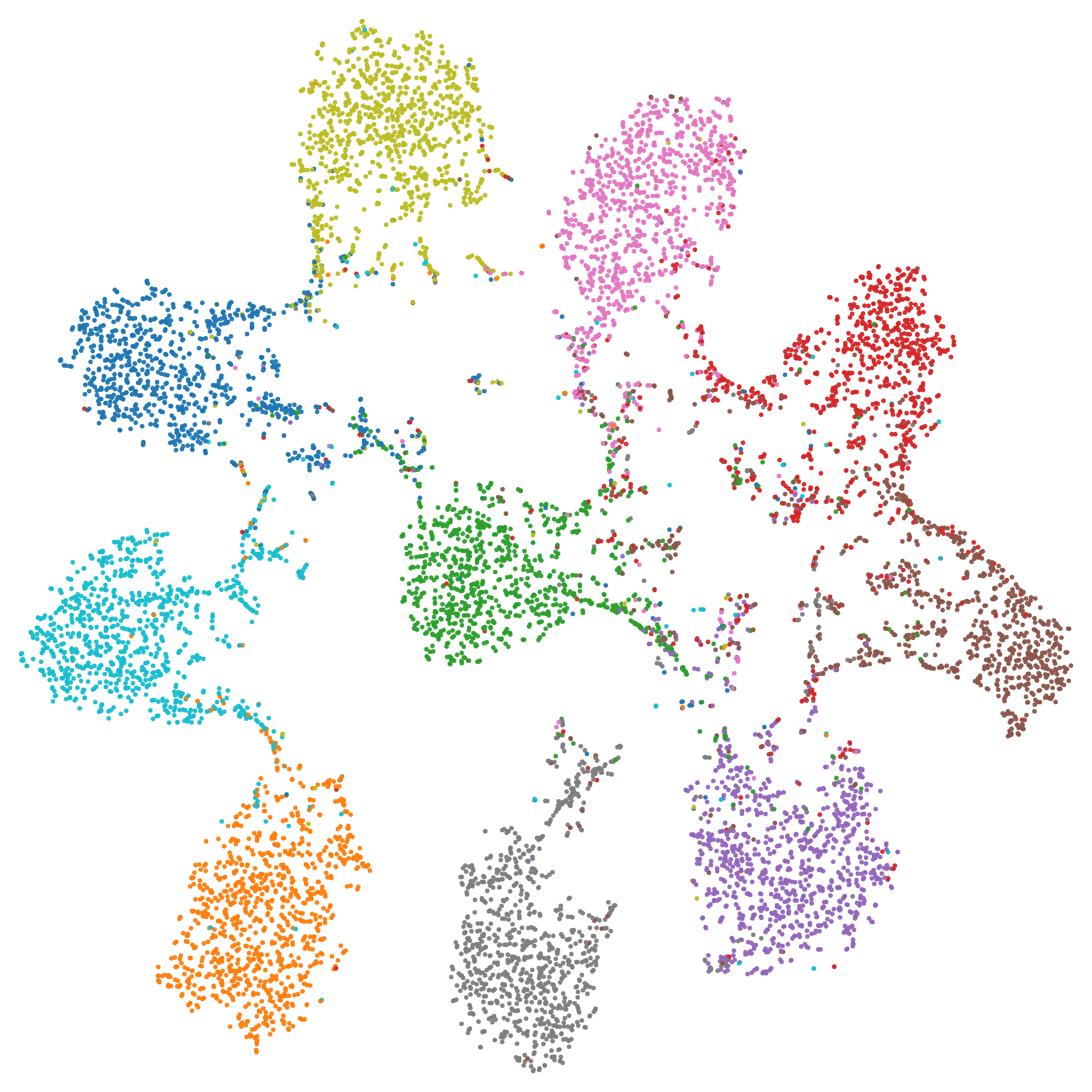}}
    \caption{CIFAR-10, Full NM-Hebb}
  \end{subfigure}
  \caption{t-SNE visualisations of ResNet-18 embeddings on CIFAR-10 after Phase~1 (a) and after the full two-phase NM-Hebb pipeline (b).}
  \label{fig:tsne-cifar10}
\end{figure*}

\begin{figure*}[htp]
  \centering
  \begin{subfigure}[t]{0.46\textwidth}
    \fbox{\includegraphics[width=\linewidth]{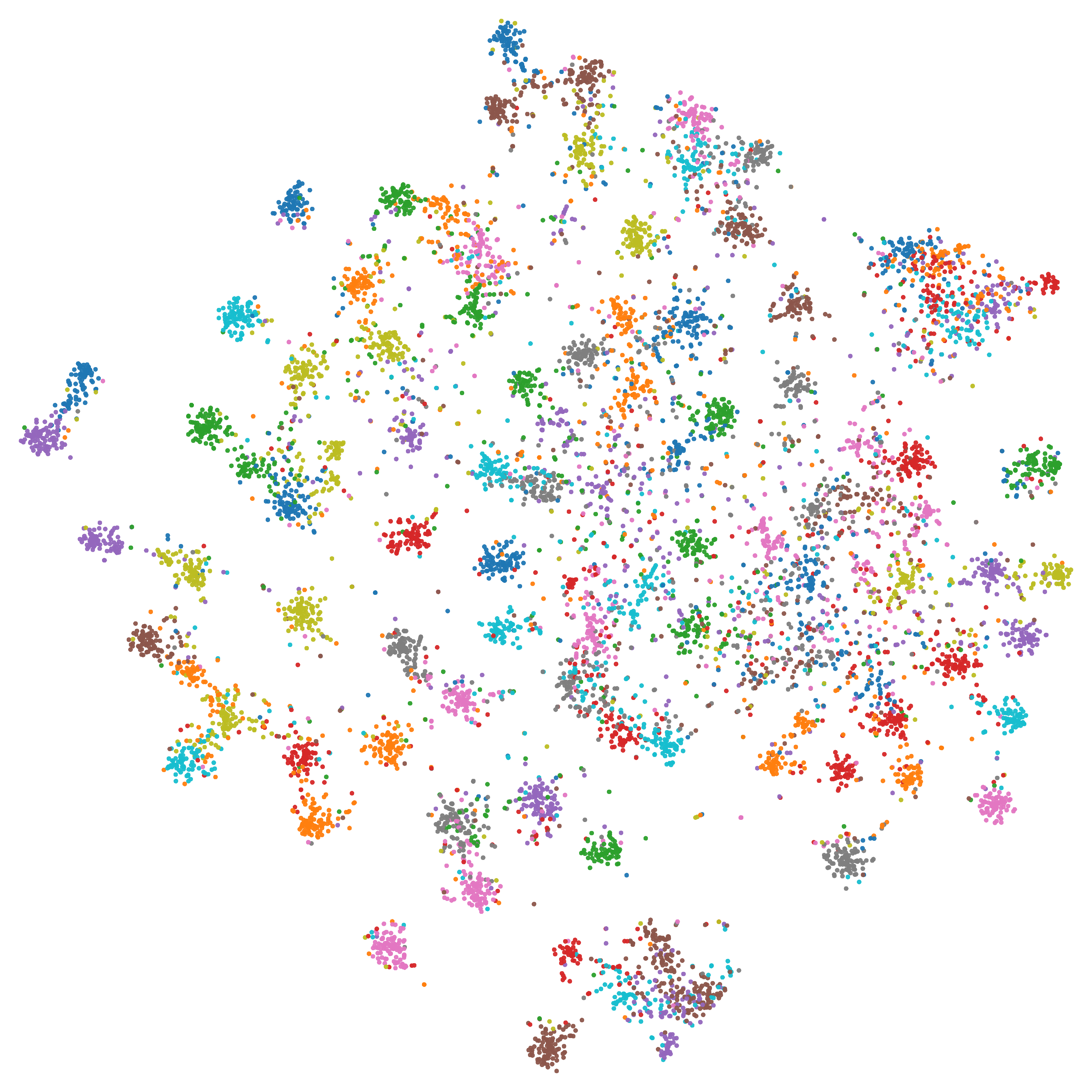}}
    \caption{CIFAR-100, Phase 1}
  \end{subfigure}\hfill
  \begin{subfigure}[t]{0.46\textwidth}
    \fbox{\includegraphics[width=\linewidth]{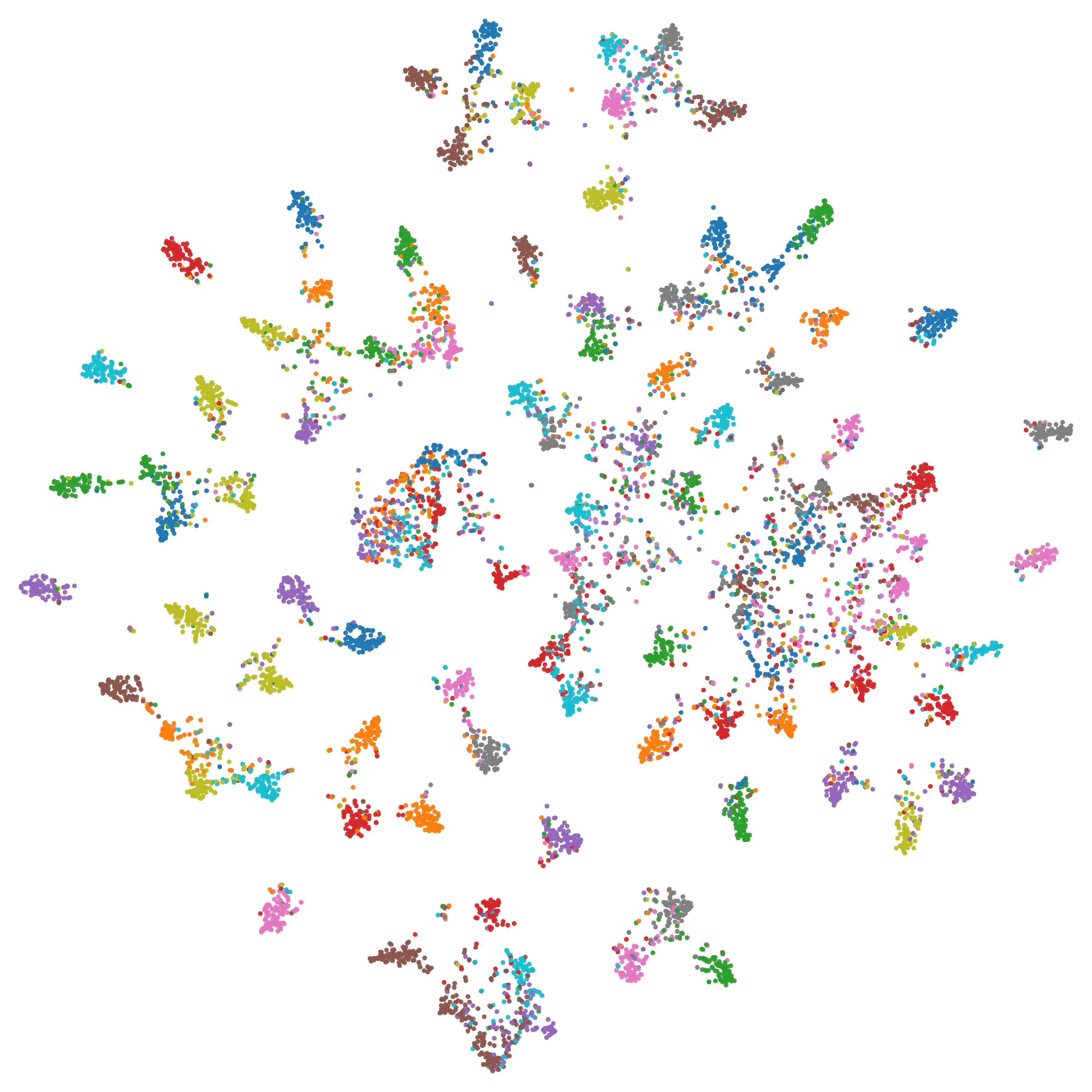}}
    \caption{CIFAR-100, Phase 2}
  \end{subfigure}
  \caption{t-SNE visualisations of ResNet-18 embeddings on CIFAR-100 after Phase~1 (a) and after the full two-phase NM-Hebb pipeline (b).}
  \label{fig:tsne-cifar100}
\end{figure*}

On CIFAR-10 (Figure~\ref{fig:tsne-cifar10}), Phase~1 already produces well-separated clusters with
limited overlap, while Phase~2 further increases inter-class margins and
reduces intra-class variance. On CIFAR-100 (Figure~\ref{fig:tsne-cifar100}), Phase~1 embeddings are
less compact due to the larger number of classes, yet Phase~2 noticeably
tightens many clusters while preserving global separation.
We omit a TinyImageNet t-SNE plot: with 200 classes, 2D projections become visually cluttered and differences are hard to read reliably.

\setlength{\fboxsep}{0pt}      
\setlength{\fboxrule}{0.5pt}   

\subsection{Filter selectivity and semantic interpretability} \label{sec:filters}
Beyond accuracy and clustering metrics, we further analyse ResNet-18 by probing filter selectivity and semantic alignment through a Network Dissection approach.

Activation selectivity (High-Activation Fraction, HAF): We quantify how often each filter produces a strong activation. Lower values indicate higher selectivity~\cite{willmore2001characterizing,haider2009neuron}. For activation selectivity, the high-activation fraction (HAF) is defined as follows.  
For a given filter $f$:
\begin{equation}
\label{eq:haf_alpha_i}
\alpha_{f}^{(i)} = \max_{u,v} a_{f}^{(i)}(u,v)
\end{equation}
where $a_{f}^{(i)}(u,v)$ denotes the activation value at spatial location $(u,v)$ in the feature map of filter $f$ for image $i$.  
Here, $\alpha_{f}^{(i)}$ is the maximum activation of filter $f$ on image $i$,  
and
\begin{equation}
\label{eq:haf_alpha_max}
\alpha_{f}^{\max} = \max_{j} \alpha_{f}^{(j)}
\end{equation}
is the maximum such activation across the entire reference set.

The high-activation fraction is:
\begin{equation}
\label{eq:haf}
\mathrm{HAF}_{f}(\tau)
= \frac{1}{N}\sum_{i=1}^{N}
\mathbf{1}\!\left\{ \alpha_{f}^{(i)} \ge \tau\,\alpha_{f}^{\max} \right\},
\qquad \tau = 0.8,
\end{equation}
where $N$ is the number of images in the set over which HAF is computed, and $\mathbf{1}\{\cdot\}$ is the indicator function. Intuitively, the indicator counts an image whenever the filter’s peak response on that image exceeds a fixed fraction $\tau$ of its own maximum recorded activation. Thus $\mathrm{HAF}_{f} \in [0,1]$ is the fraction of images on which filter $f$ fires strongly.  
Lower HAF values indicate more selective filters, while values closer to $1$ indicate broad, non-selective responses.

In parallel, we quantify a speckle rate—the proportion of filters whose spatial structure is noise-like rather than oriented or textured. Operationally, we zero-mean/unit-variance normalise each kernel, compute its 2-D Fourier magnitude, and label it as speckle if (i) at least 60\% of the total power lies beyond half the maximum radius (outer 50\% of the radial spectrum; high-frequency dominance) and (ii) the periodic (modulo $\pi$) Fourier-domain orientation histogram has low concentration (resultant length $\le$ 0.20). This frequency-orientation criterion follows standard practice in frequency/texture analysis and directional statistics \cite{field1987relations,portilla2000parametric,bruna2013invariant} and aligns with evidence that CNNs are sensitive to frequency content and texture bias \cite{geirhos2020shortcut,yin2019fourier}.

\begin{table}[ht]
\caption{Filter morphology statistics for the Hebbian-regularised layer in ResNet-18 (median Top-1 run across 5 seeds). Lower HAF indicates higher selectivity.}
\label{tab:filter_morphology}
\small
\centering
\begin{tabular}{|l|c|c|c|c|c|c|}
\hline
\textbf{Training regime} & \multicolumn{2}{c|}{\textbf{CIFAR-10}} & \multicolumn{2}{c|}{\textbf{CIFAR-100}} & \multicolumn{2}{c|}{\textbf{TinyImageNet}} \\
\cline{2-7}
 & Speckle (\%) & HAF ($\downarrow$) & Speckle (\%) & HAF ($\downarrow$) & Speckle (\%) & HAF ($\downarrow$) \\
\hline
Baseline         & 64 & 0.480 & 67 & 0.520 & 9   & 0.38    \\
Phase 1 NM-Hebb  & 28 & 0.271 & 22 & 0.205 & 3.96 & 0.226 \\
Full NM-Hebb  & 18 & 0.189 & 17 & 0.195 & 1.56 & 0.197 \\
\hline
\end{tabular}
\end{table}

\smallskip
Furthermore, we measure the alignment of filters with human-interpretable concepts (edges, textures, objects, colors) using the Network Dissection framework~\cite{bau2017network,olah2018building}. This provides a high-level view of interpretability that complements selectivity metrics.

To jointly visualise selectivity and semantic alignment across training regimes, we apply a Network Dissection probe~\cite{bau2017network} to the final convolutional block of ResNet-18 (512 filters, $3{\times}3{\times}512$). For each filter, we compute the pixelwise intersection-over-union (IoU) between thresholded activation maps (top $0.5\%$ of activations) and Broden semantic masks, using Broden-224 as input. Since these backbones were trained on relatively low-resolution datasets (CIFAR-10/100 and TinyImageNet), we restrict our analysis to the texture category, as other Broden families yielded negligible or sub-threshold coverage.

We quantify interpretability by counting the number of filters that surpass the standard
Network Dissection threshold ($\text{IoU} \geq 0.05$)~\cite{bau2017network}.
As shown in Table~\ref{tab:semantic_filters}, the baseline models yield only a small proportion of texture-selective units (10–15\%). By contrast, color-selective filters never emerge on CIFAR datasets: their maximum IoU remains consistently below 0.02. TinyImageNet reaches somewhat higher scores (up to $\sim 0.04$), but still falls short of the semantic threshold ($\text{IoU} \geq 0.05$). This pattern is consistent with the limited chromatic diversity of CIFAR-10/100 and, more broadly, with prior reports that small CNNs prioritize structural over chromatic cues~\cite{zhang2018interpretable,neyshabur2020towards}. Phase~1 NM-Hebb substantially increases texture alignment across all three datasets: from 285 filters (55.4\%) on CIFAR-10, to 185 (36.1\%) on CIFAR-100, and 210 (41.0\%) on TinyImageNet. Full NM-Hebb further consolidates this effect.
Interestingly, while CIFAR models remain almost exclusively texture-driven, TinyImageNet exhibits a broader distribution of near-threshold responses ($0.04 \leq \text{IoU} < 0.05$) in non-texture families such as parts, objects, colors, and scenes. Although these do not cross the semantic cutoff, they suggest richer but weaker alignment, with a steadier trend in the number of activated filters — consistent with TinyImageNet’s closer match to Broden-224 content.

\begin{table}[ht]
\caption{Number of texture-selective filters ($\text{IoU} \geq 0.05$) in the final convolutional block of ResNet-18 (512 filters total).}
\label{tab:semantic_filters}
\small
\centering
\begin{tabular}{|l|c|c|c|}
\hline
\textbf{Training regime} & \textbf{CIFAR-10} & \textbf{CIFAR-100} & \textbf{TinyImageNet} \\
\hline
Baseline        & 74 \;(14.8\%)   & 58 \;(11.3\%)  & 65 \;(12.7\%)  \\
Phase~1 NM-Hebb & 285 \;(55.4\%)  & 185 \;(36.1\%) & 210 \;(41.0\%) \\
Full NM-Hebb    & 314 \;(61.3\%)  & 201 \;(39.3\%) & 225 \;(43.9\%) \\
\hline
\end{tabular}
\end{table}

Overall, neuromodulated Hebbian updates substantially expand the pool of texture-selective filters, stabilizing low-level structural features and yielding a more interpretable representation without additional supervision.
\section{Conclusion}\label{sec:concl}
This work introduced NM-Hebb, a two-phase training procedure
that integrates Hebbian regularisation, neuromodulated consolidation and
pairwise metric shaping into a conventional SGD pipeline.
Across five convolutional backbones (3.5–11 M parameters) the approach
yielded absolute improvements of +2–10 pp in Top-1
accuracy and up to 0.15 in NMI on CIFAR-10 and CIFAR-100, relative to
standard supervised learning based on cross-entropy loss, as well as techniques such as knowledge distillation and contrastive learning.
When extended to TinyImageNet, training from
scratch without ImageNet pretraining for 120 epochs, NM-Hebb delivers
consistent but smaller gains of +4.3–8.9 pp in Top-1 accuracy
across the same backbones, accompanied by concomitant increases in
embedding compactness (NMI). Visual analyses of embeddings and
convolutional filters across all three datasets indicate cleaner class
separation and more structured low-level primitives, achieved with a
modest increase in training time. In addition to accuracy and clustering improvements, NM-Hebb consistently enhanced filter selectivity and semantic alignment, providing a more interpretable representation space alongside better classification performance.

The evaluation focused on small-resolution benchmarks (CIFAR and TinyImageNet) and single-GPU training; large-scale
datasets, heavy class imbalance and multi-modal inputs were not
considered. In addition, the neuromodulator was limited to a two-layer
perceptron; alternative gating mechanisms or event-driven
implementations remain to be explored.

Future work will extend NM-Hebb to large-scale datasets (e.g., ImageNet-1k) and hybrid transformer–convolution architectures, assess robustness to distribution shifts and adversarial perturbations, explore continual learning without explicit task boundaries, and investigate low-precision or neuromorphic hardware for on-device adaptation.

\section*{CRediT authorship contribution statement}
\textbf{Davorin Miličević}: Conceptualization, Data curation, Methodology, Software, Investigation, Visualization, Writing – Original Draft, Writing – Review \& Editing.  
\textbf{Ratko Grbić}: Conceptualization, Supervision, Methodology, Validation, Resources, Writing – Review \& Editing.

\section*{Declaration of generative AI and AI-assisted technologies in the writing process}
During the preparation of this work the author(s) used ChatGPT in order to improve language and readability. After using this tool/service, the author(s) reviewed and edited the content as needed and take(s) full responsibility for the content of the publication.
\bibliographystyle{elsarticle-num}

\bibliography{refs}  
\end{document}